%% file: main.tex
\documentclass[10pt,twocolumn,letterpaper]{article}

\usepackage{cvpr}              


\definecolor{cvprblue}{rgb}{0.21,0.49,0.74}
\usepackage[pagebackref,breaklinks,colorlinks,allcolors=cvprblue]{hyperref}
\usepackage{tabularx}
\usepackage{bigfoot}
\usepackage[flushleft]{threeparttable}
\usepackage{hyperref} 
\def\paperID{*****} 
\def\confName{CVPR}
\def\confYear{2025}

\title{PhytoSynth: Leveraging Multi-modal Generative Models for Crop Disease Data Generation with Novel Benchmarking and Prompt Engineering Approach}

\author{Nitin Rai\textsuperscript{1}, \hspace{0.5pt} Arnold Walter Schumann\textsuperscript{2}\thanks{Corresponding author}, \hspace{0.5pt} Nathan Boyd\textsuperscript{1} \\
\textsuperscript{1}Gulf Coast Research and Education Center (GCREC), University of Florida, Wimauma, FL, USA \vspace{-1.65mm}
\and
\textsuperscript{2}Citrus Research and Education Center (CREC), University of Florida, Lake Alfred, FL, USA \\
{\tt\small \textbraceleft \href{mailto:nitin.rai@ufl.edu}{nitin.rai}, \href{mailto:schumaw@ufl.edu}{schumaw}, \href{mailto:nsboyd@ufl.edu}{nsboyd}\textbraceright{@ufl.edu}}
}

\begin{document}
\maketitle
\input{sec/0_abstract}    
\input{sec/1_intro}

\input{sec/2_related_works}
\input{sec/3_proposed_dataset}
\input{sec/4_mandm}
\input{sec/5_results}
\input{sec/6_discussion}
\input{sec/7_conclusions}
\input{sec/8_ackn}
{
    \small
    \bibliographystyle{ieeenat_fullname}
    \bibliography{main}
}


\end{document}

%% file: sec/0_abstract.tex
\begin{abstract}
Collecting large-scale crop disease images in the field is labor-intensive and time-consuming.~Generative models (GMs) offer an alternative by creating synthetic samples that resemble real-world images.~However, existing research primarily relies on Generative Adversarial Networks (GANs)-based image-to-image translation and lack a comprehensive analysis of computational requirements in agriculture.~Therefore, this research explores a multi-modal text-to-image approach for generating synthetic crop disease images and is the first to provide computational benchmarking in this context.~We trained three Stable Diffusion (SD) variants—SDXL, SD3.5M (medium), and SD3.5L (large)-and fine-tuned them using Dreambooth and Low-Rank Adaptation (LoRA) fine-tuning techniques to enhance generalization.~SD3.5M outperformed the others, with an average memory usage of 18 GB, power consumption of 180 W, and total energy use of 1.02 kWh/500 images ($\approx$0.002 kWh/image) during inference task.~Our results demonstrate SD3.5M's ability to generate 500 synthetic images from just 36 in-field samples in $\approx$ 1.5 hours. We recommend SD3.5M for efficient crop disease data generation.
\end{abstract}

%% file: sec/1_intro.tex
\section{Introduction}
\label{sec:intro}
Present agricultural robotic technologies rely heavily on computer vision algorithms for smart decision-making.~However, to develop reliable and adaptable computer vision algorithms, a diverse set of very large-scale datasets are needed.~The current practice of acquiring large-scale datasets in agriculture is primarily based on making multiple field trips during the growing season, which can be a time-consuming and labor-intensive procedure~\cite{heider2025survey, lu2020survey}. Moreover, even if the datasets are made open access, addressing the specific need of the user--such as dynamic environmental conditions and multiple crop growth stages, is a bottleneck.~Thus, alternative methods for automating data generation from small sample collections are necessary.

Generative artificial intelligence (Gen AI) offers a promising approach for creating large-scale datasets from a very few samples. As a subfield of AI, it can mimic real-world images to generate synthetic data by capturing fine-grained features representative of specific classes. However, the use of multi-modal generative models (GMs) for image generation remains underexplored, particularly in dynamic crop disease context.~\textit{Therefore, in this research, we employ a multi-modal GM to generate synthetic images of common diseases in watermelon crops.}~Unlike weed datasets~\cite{deng2025weed,moreno2023analysis,pathak2023machine}, crop disease datasets are more complex due to their variability in symptoms and spatial-temporal diversity, making it challenging for GMs to accurately replicate disease patterns. 

Most disease-centered images in agriculture are trained using Generative Adversarial Networks (GANs) with image-to-image translation~\cite{lu2022generative, jin2022grapegan, lu2020survey}.~\textit{Furthermore, no study has comprehensively addressed critical aspects of training a GM for disease-related applications, such as memory consumption, power usage, energy dissipation, and time required to generate a desired number of samples}.~These factors are crucial in agricultural contexts, as understanding can help answer key questions:~(a) What strategies should be followed to generate realistic samples from disease datasets?~(b) How does prompt engineering play a vital role in generating highly realistic disease images?~(c) How can adopting a cost-effective approach in training a GM help small organizations with limited access to high-performance computing?~(d) How can flexibility and speed in the generation of synthetic disease samples ensure that AI-driven technologies are capable of addressing pressing on-farm disease threats?~(e) How can the creation of a standardized benchmarking framework help users and researchers make informed decisions when selecting a GM for synthetic disease generation?

Considering the aforementioned gaps in multiple aspects, this research study aims to address these questions by leveraging Stable Diffusion (SD) models based on text-to-image generation for crop disease image creation. In addition, \textit{we are the first research to also report novel benchmarking for computational requirements and prompt engineering strategies needed to generate high quality and relevant crop disease dataset}. Overall, Sec.~\ref{sec:relatedwork} discusses related work and the need for SD models in agriculture, Sec.~\ref{sec:proposeddata} highlights dataset used and strategies to create a good training set, Sec.~\ref{mandm} explores the inner workings of SD models when trained on crop disease dataset along with Dreambooth and Low-Rank Adaptation (LoRA) finetuning approaches, Sec.~\ref{results} reports relevant metrics, and finally, Sec.~\ref{discussion} \& \ref{conclude} present findings and observations, and conclusions, respectively.~The key contributions are:

\begin{enumerate}
    \item Train and fine-tune multiple variants of the multimodal SD models using Dreambooth and LoRA techniques to generate realistic synthetic crop disease samples.
    \item Measure and report the computational resources (e.g., GPU memory, power consumption, and energy) required to train and generate synthetic crop disease images.
    \item Establish benchmarking and standardization to determine the best model for generating crop disease images.
\end{enumerate}

%% file: sec/2_related_works.tex
\section{Related Work}
\label{sec:relatedwork}
\subsection{Generative adversarial network (GAN) models}
With advances in synthetic data generation, GAN-based models have proven to be beneficial for agriculture.~Many studies have explored various GAN variants with Cycle-Consistent (CycleGAN) and Deep Convolutional (DCGAN) being the most widely used for generating synthetic samples. For example, Yunong et al.~\cite{tian2019detection} trained CycleGAN to perform data augmentation by generating diseased samples of apple fruit. Similarly,  Asenovic et al.~\cite{arsenovic2019solving} trained a DCGAN to generate synthetic data of multiple crop diseases. Another research by Wu et al.~\cite{wu2020dcgan} demonstrated the applicability of generating tomato leaf diseases using DCGAN models. 

In general, a comprehensive review study by Lu et al.~\cite{lu2020survey} examined the application of GAN architectures for disease data generation, showcasing the use of more than 15 different variants to generate synthetic samples of various crop diseases.~\textit{However, GAN architectures have been found to pose several challenges in areas such as training instability~\cite{goodfellow2020generative}, generating low-quality images~\cite{zhu2020data}, and demand for large volumes of training set~\cite{NIPS2016_8a3363ab}.~Based on the study by~\cite{lu2020survey}, it is clearly reported that although GAN is the most adopted approach to generate synergetic image data, it has limitations in various areas. Therefore, this research does not rely solely on using GANs but uses state-of-the-art (SoTA) diffusion models for text-to-image generation.}

\subsection{Diffusion models}

With the growing interest in using diffusion models to generate high-quality images in the medical field~\cite{kazerouni2023diffusion}, product generation~\cite{li2023planning}, and urban applications~\cite{reutov2023generating}, application in the agricultural domain remains underexplored~\cite{muhammad2023harnessing}.~Although recent studies have explored the potential to leverage diffusion architectures to generate high-quality synthetic samples of crop diseases, no research has demonstrated its comprehensive benchmarking in generating synthetic samples.~For example, a study by Mori et al.~\cite{mori2024application} proposed using a latent diffusion model to generate images of diseased apple leaves employing an image-to-image generation approach.~Another study by Zhou et al.~\cite{zhou2025novel} addressed data scarcity by integrating the diffusion model with a few-shot learning technique to implement an end-to-end pipeline for plant disease identification.~A regression-conditional based image-to-image diffusion model was developed by Egusquiza et al.~\cite{egusquiza4886401synthetic} to create graded synthetic data to quantify plant disease severity. 

Based on the limited studies that have explored the potential of multimodal GM to generate synthetic disease images, the work of Muhammad et al.~\cite{muhammad2023harnessing} is the only work that has reported a comparative study between the RePaint diffusion model and InstaGAN. The authors have evaluated the effectiveness of diffusion-based generated images using the image-to-image translation approach rather than text-to-image generation. Moreover, the text-to-image methodology allows more control over the created samples by imparting variations in the synthetic samples compared to the image-to-image approach~\cite{saharia2022photorealistic}.~\textit{In addition to this, similar study by Muhammad et al.~\cite{muhammad2023harnessing} does not address model comparison in terms of GPU power usage and various other aspects related to training and ease of image generation.~Furthermore, there are no standardized or universal computational metrics that can evaluate the performance of Gen-AI models in generating synthetic images specifically targeting crop diseases~\cite{lu2020survey}.~Therefore, the benchmarks and computational metrics proposed in this study address this limitation, with the aim of advancing more efficient synthetic data generation in the agricultural scenario.} 

%% file: sec/3_proposed_dataset.tex
\section{Proposed dataset}
\label{sec:proposeddata}

\subsection{Our dataset}

\begin{figure*}[htbp]
  \centering
    \includegraphics[scale=0.52]{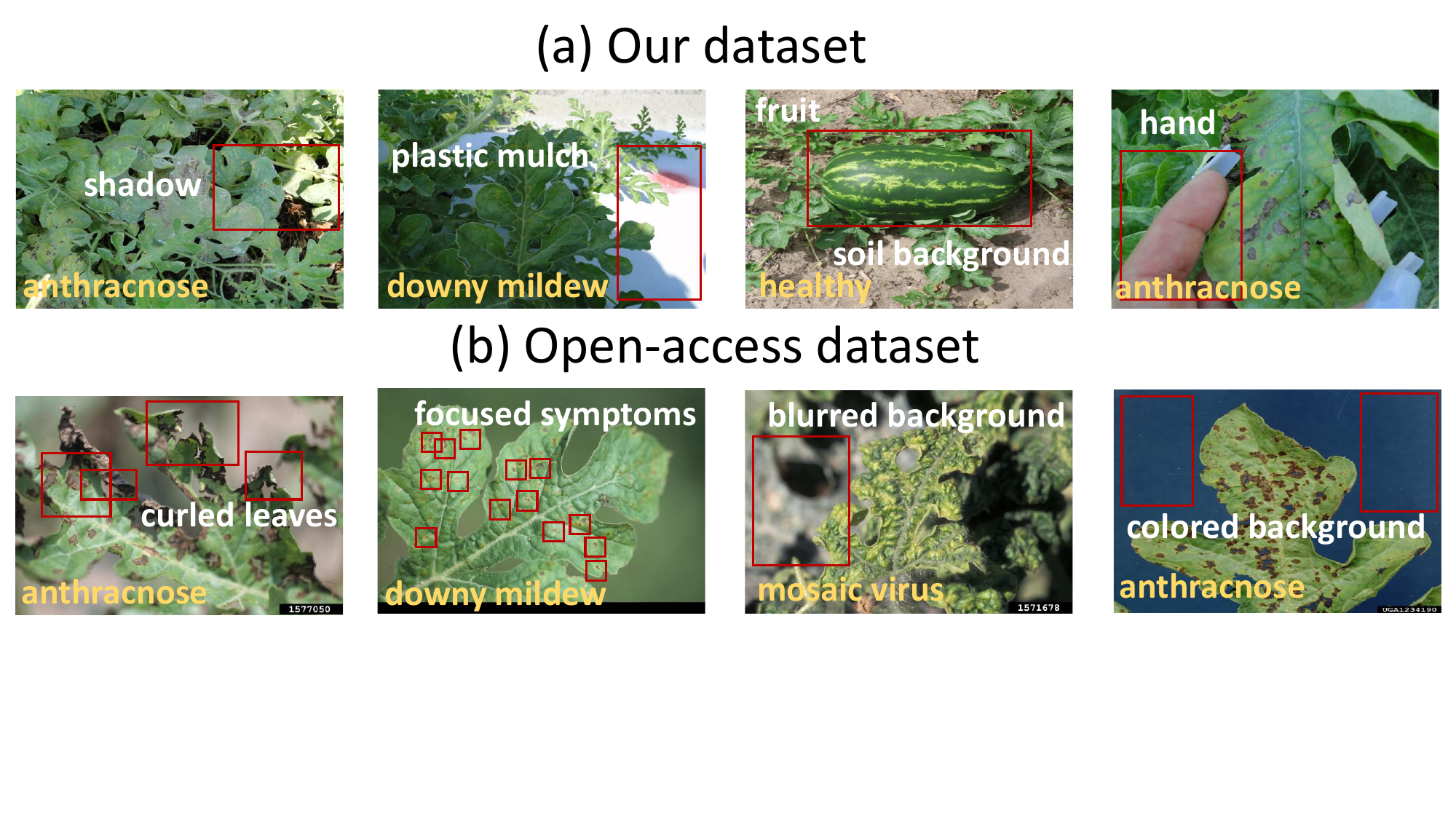}
    \vspace{-30mm}
    \caption{Overview of proposed dataset for this research study: (a) our dataset that consists of variations, such as shadow, plastic mulch, fruit, soil background, and a hand, and (b) open-access dataset that imparts additional diversity in terms of curled leaves, focused symptoms, blurred and colored backgrounds.}
    \label{figdataset}
\end{figure*}

To generate images, we selected common disease symptoms that are caused by fungal infections in watermelons \textit{(Citrullus lanatus)}, specifically anthracnose (AT) \textit{(Colletotrichum orbiculare)} and downy mildew (DM) \textit{(Pseudoperonospora cubensis)}. The input dataset used for training the GMs was sourced from two origins.~The first source was our own dataset, for which we carried out fieldwork to capture images of disease symptoms in watermelons~(Fig.~\ref{figdataset}a). This approach allowed us to selectively capture samples from the field with desired variations and orientations (Fig.~\ref{figdataset}a).~Images for AT and DM were captured using Sony's handheld digital single-lens reflex (DSLR) camera.~The time frame for data acquisition was from the 5\textsuperscript{th} of May until 15\textsuperscript{th} June.~All the images were collected at a high resolution of 2048 $\times$ 1536.~In total, we gathered 472 images of AT and 45 of DM. Following this, we implemented multiple filtering steps to ensure that each symptom was a best representation of a specific class. For instance, in Table~\ref{tab1}, only 12 out of 472 images were selected for AT as the remaining images did not clearly represent the disease symptoms.~This step was further improved with the help of an expert in the domain of plant pathology.~See Table~\ref{tab1} for the overall final image count after discarding those with inconsistent symptoms for their assigned class.
\subsection{Open-access dataset}
We downloaded a few more samples from the \href{https://www.ipmimages.org/}{\textcolor{blue}{Integrated Pest Management (IPM)}}\footnote{Open-access images were captured by:~\href{https://strawberry.calpoly.edu/faculty-staff}{\textcolor{blue}{G. J. Holmes.}}} dataset repository.~Additional downloaded datasets for AT and DM were also added to the set. We purposely mixed our dataset with open access images because it imparted greater variation in disease expression, such as curled leaves, focused symptoms, blurred and colored backgrounds (Fig.~\ref{figdataset}b).~This approach improves data quality, covers multiple crop growth stages, and adds variations in terms of background complexities~\cite{phan2022identification}.~The open access images from the IPM repository were downloaded at multiple resolutions, which were later converted to 1024 $\times$ 1024 resolution for model training.
\begin{table}[h]
\footnotesize
\centering
\begin{threeparttable}
  \begin{tabularx}{1\linewidth}{X c c c}
    \toprule
    Class & Original set & Filtered out & Final training set\\
    \midrule
    \multicolumn{4}{c}{\textbf{Our dataset}} \\
    \midrule
    Anthracnose\textsuperscript{\textcolor{blue}{$\dagger$}} & 472 & 460 & 12 \\ 
    Downy mildew & 45 & 30 & 15 \\
    \bottomrule
    \multicolumn{4}{c}{\textbf{Open-access dataset}} \\
    \midrule
    Anthracnose\textsuperscript{\textcolor{red}{$\ddagger$}} & 32 & 8 & 24 \\
    Downy mildew & 45 & 37 & 8 \\
    \bottomrule
  \end{tabularx}
\caption{Overview of disease dataset samples for training the models. \textit{\textbf{Note:}} We have reported results based on~\textcolor{blue}{$\dagger$} + \textcolor{red}{$\ddagger$} (36 images) due to space constraint.~For other classes, results were similar.}
\label{tab1}
\end{threeparttable}
\end{table}

\subsection{Overall dataset considerations and strategies}
After filtering, all images were exported to~\textit{*.png} format to preserve important features and resized to 1024 $\times$ 1024 resolution.~A key consideration when assembling disease data for each class was maintaining consistency, as mixing different views (e.g, close-up with a hand-held camera and aerial with drones) within the training set led to poor results.~Another strategy to include while training the model is not to include multiple disease symptoms in the training set.~For instance, disease samples of AT and DM should not be mixed while training the model as it will create unrealistic disease symptoms.~Overall, training data should be tailored to the specific application.~Additionally, while some fungal infections, such as downy mildew (DM) or powdery mildew \textit{(Podosphaera xanthii)}, can affect multiple crops within the cucurbit family, mixing images of cucumbers and watermelons may generate unrealistic images.

%% file: sec/4_mandm.tex
\section{Methodology}
\label{mandm}
\subsection{Overview \& experimental setup} \label{overview setup}

\begin{figure}[htbp]
    \includegraphics[width=1\columnwidth]{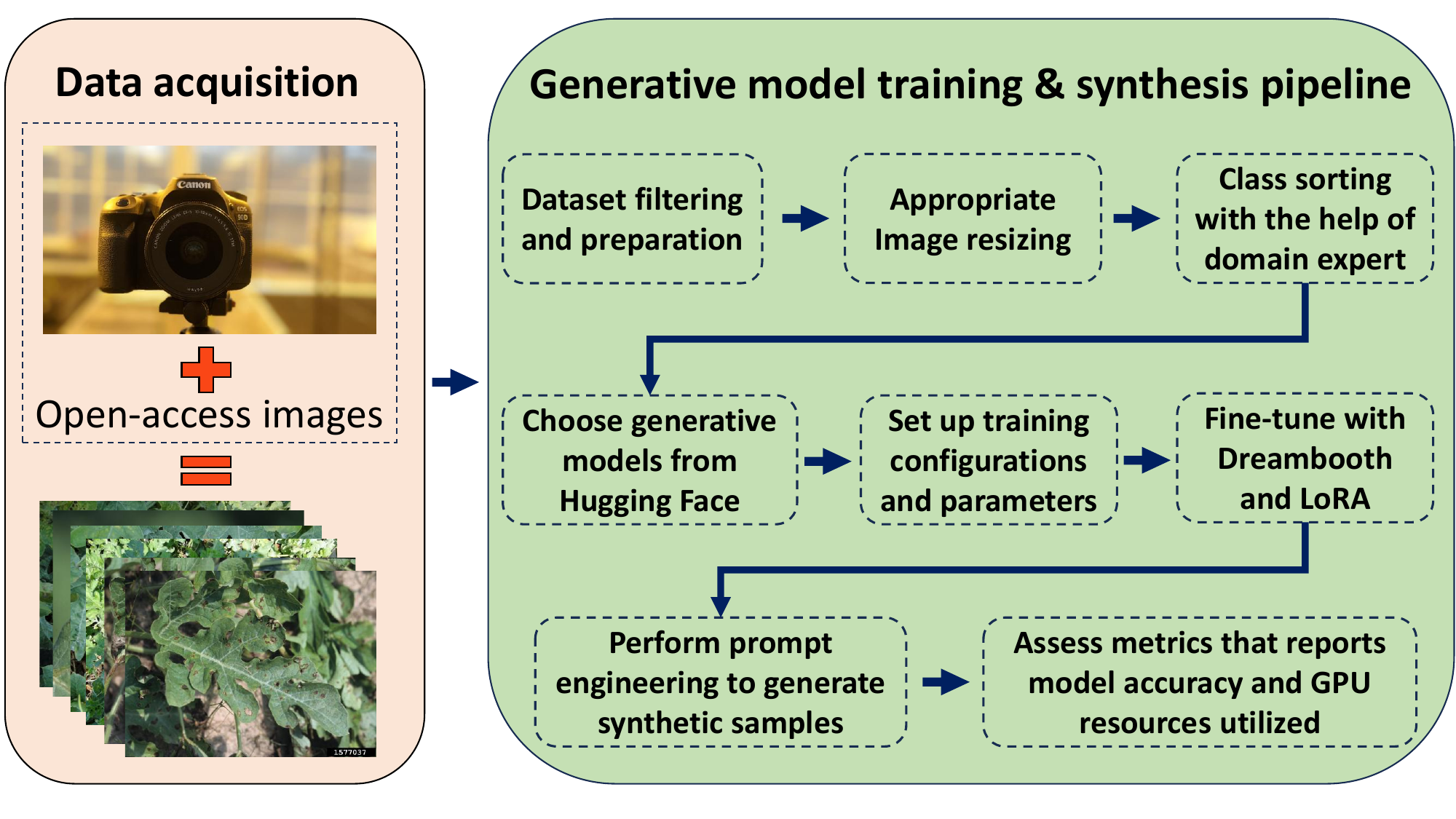}
    \caption{Pipeline for generative model training and synthetic image generation. \textbf{\textcolor{Apricot}{Left box:}} The workflow begins with data acquisition, incorporating both captured and open-access images.~\textbf{\textcolor{LimeGreen}{Right box:}} The acquired images undergo dataset filtering, resizing, and class sorting to train a generative model using Dreambooth and LoRA fine-tuning approaches. Prompt engineering is employed to generate user-centered desired number of synthetic images.}
    \label{figoverallmethod}
\end{figure}

This section describes the overall methodological pipeline and platform used to train and generate disease-centered synthetic images (Fig.~\ref{figoverallmethod}).~Furthermore, we demonstrate that SD models are modifiable and could be fine-tuned through Dreambooth~\cite{ruiz2023dreambooth} and LoRA~\cite{liu2024lora}.~We also discuss the inner workings of SD models and explain how Dreambooth and LoRA enhance the learning of disease-specific features, resulting generation of high-quality synthetic images.

We present the model metrics in the subsequent sections, focusing on assessing computational requirements.~The overall analysis was carried out on a high performance computing system (HPC)~(\href{https://www.rc.ufl.edu/about/hipergator/}{\textcolor{blue}{HiPerGator}}) equipped with a single Nvidia A100 Quadro with 70 GB of VRAM, 64 GB of RAM, and 10 CPU cores.~The codes were executed on the Jupyter Notebook platform using Python (3.11.11), Torchvision (0.20.0), Torch (2.5.1), and Transformers (4.47.0), with CUDA 12.4 support and \href{https://github.com/huggingface/diffusers}{\textcolor{blue}{Hugging Face}}.

\subsection{Stable diffusion model architecture}
The SD models are advancing synthetic image generation in agriculture.~\textit{To our knowledge, SD architectures using text-to-image prompts have not been explored for crop disease dataset generation.}~Therefore, this research trains three SD variants--SDXL, SD3.5M (medium), and SD3.5L (large). Beyond generating synthetic samples, we enhance our approach by integrating Dreambooth~\cite{ruiz2023dreambooth} and LoRA~\cite{hu2022lora} fine-tuning.~Overall, SDXL has 3.5 bn parameters, while SD3.5M and SD3.5L contains 2.5 and 8 bn, respectively.
\begin{figure*}
    \includegraphics[scale=0.52, trim={0 5cm 0 0}, clip]{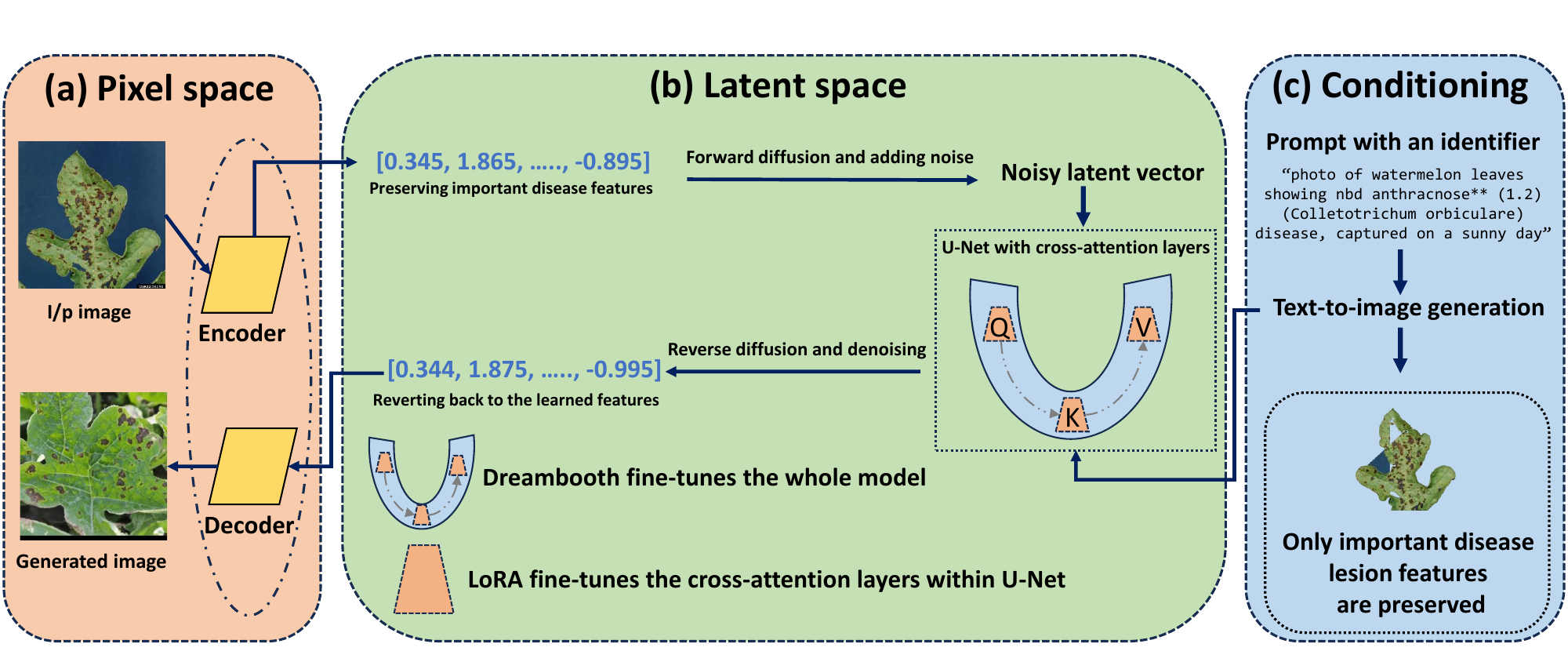}
    \caption{Schematic representation of the generative pipeline for disease image synthesis using a stable diffusion model: (a) In the pixel space, input images are encoded and generated images are decoded after training, (b) In the latent space, the model undergoes forward (noising process) and reverse (denoising process) diffusion process, and (c) The conditioning stage enables text-to-image generation using descriptive prompts while ensuring relevant disease features are preserved.}
    \label{figSD}
  \end{figure*}
\subsubsection{Model explainability (xAI) for agricultural dataset} \label{modelexplain}
AI model explainability (xAI) is a major research gap in agriculture, often labeled a black box approach~\cite{saranya2023systematic, ryo2022explainable}.~As researchers increasingly use SD models to generate synthetic samples for their custom applications, we explore their its inner working for disease data generation.

The SD architecture consists of three major components: pixel space, latent space, and semantic conditioning (Fig.~\ref{figSD}). It takes two inputs--images and prompts (Fig.~\ref{figSD}a, c)--which are converted into vector embeddings via a convolutional neural network and a transformer model, respectively.~In addition to this, Contrastive Language-Image Pre-training (CLIP)~\cite{radford2021learning} generates embeddings for both inputs within the same latent space (Fig.~\ref{figSD}b).~In the pixels-space component, images pass through an encoder (Fig.~\ref{figSD}a), converting them into latent representations (Fig.~\ref{figSD}b).~This reduces the computational load while preserving key disease features like lesion shape and texture.~In the latent space, the model applies noise through forward diffusion and learns to remove it via reverse diffusion, thereby learning image distribution in the process by transforming structured image into noisy one.~This process is driven by a U-Net architecture with cross-attention layers.~In the conditioning component (Fig.~\ref{figSD}c), the transformer model guides prompts toward semantic disease mapping.~For instance, given the prompt,~\texttt{\textcolor{BrickRed}{``~a photo of watermelon showing nbd anthracnose\textsuperscript{**} (1.2) (Colletotrichum orbiculare) disease''}}, U-Net gradually denoises the image to best represent the disease symptoms.~Once denoising is complete, the decoder reconstructs and upsamples the image back to pixel space (Fig.~\ref{figSD}a). 
\subsubsection{Dreambooth for fine-tuning whole model}
Dreambooth helps to fine-tune the whole model mainly by aiming to modify the denoising process in the latent space~\cite{ruiz2023dreambooth}. As the U-Net model learns to predict the noise in the given latent based on the actual noise, Dreambooth minimizes the variation between the actual and predicted noise. During the forward diffusion process, noisy images $(x\textsubscript{t})$ is generated by adding Gaussian noise on the input image. It is given by the equation (Eq.~\ref{equ1}). 

\begin{equation}
    x_t=\sqrt{\alpha_t}x_0 + \sqrt{1-\alpha_t}\epsilon, \text{ } \epsilon\sim N(0, I)
    \label{equ1}
\end{equation}

\noindent where, $x_t$ is the noisy image, $x_0$ is the original image, $\alpha_t$ is the noise schedule parameter, and $\epsilon$ is random Gaussian noise~\cite{nikolov2023general}. Furthermore, during the reverse denoising process (Fig.~\ref{figSD}b), the model learns a denoising function ($\epsilon_\theta$, $x_t$, $p_t$) parameterized by $\theta$ trained to predict noise ($\epsilon$) in the given $x_t$ and prompt ($p_t$), given by Eq.~\ref{equ2}.~Therefore, during fine-tuning, Dreambooth optimizes $\hat{\epsilon_\theta}$ equal to $\epsilon$ so that it can predict and remove noise and reconstruct better synthetic images. Overall, the ultimate objective of Dreambooth is to minimize the loss function, i.e., mean squared error between the actual and predicted noisy images which is given in Eq.~\ref{equ3}, where $L(\theta)$ is the loss function and $\mathbb{E}\textsubscript{x\textsubscript{0,t,$\epsilon$}}$ are the expected values of the aforementioned variables.

\begin{equation}
    \hat{\epsilon_\theta} = \epsilon_\theta(x_t, t, p_t)
    \label{equ2}
\end{equation}
\begin{equation}
    L(\theta) = \mathbb{E}\textsubscript{x\textsubscript{0,t,$\epsilon$}}[||\epsilon_\theta(x_t, t, p_t)-\epsilon||^2_2]
    \label{equ3}
\end{equation}

\subsubsection{LoRA for cross-attention layers}

LoRA fine-tuning is specifically applied to the self-attention layers Q (query), K (key), and V (value) within the U-Net architecture (see Fig.~\ref{figSD}b)~\cite{hu2022lora}. The main objective of LoRA is to introduce low-rank matrices within these layers (without updating the full weight parameters) to reduce computational complexity while still enhancing the model to adapt to new concepts. It updates the attention layers according to Eq.~\ref{eq4}, where $\Delta W_Q$ equals $A_QB_Q$ with $A_Q$ $\in$ $\mathbb{R}\textsuperscript{d$\times$r}$, $B_Q$ $\in$ $\mathbb{R}\textsuperscript{r$\times$d}$, r is a low-rank decomposed matrix and d is original ranked matrix.~Similar fine-tuning through LoRA is followed with other attention layers, K and V. \vspace{-0.1cm}

\begin{equation}
    W'_Q = W_Q + \Delta W_Q
    \label{eq4}
\end{equation}

\subsection{Strategies: Prompts and hyperparameters}

Prompt engineering and hyperparameter tuning approaches were integrated to generate diverse and synthetic images that relates closely to the training set (Table~\ref{tab2}).~Prompt engineering strategies are clubbed in two ways:~(a) associating a unique key identifier in the prompt, and (b) adding weights on the key features that need to be highlighted in the generated images.~Based on Fig.~\ref{figSD}c, the prompt,~\texttt{\textcolor{BrickRed}{``a photo of watermelon showing nbd anthracnose\textsuperscript{**} (1.2) (Colletotrichum orbiculare) disease''}}, incorporates the key identifier~\texttt{\textcolor{BrickRed}{``nbd''}} to distinctly associate the generated image with anthracnose disease.~This is a very novel approach that helps with fine-tuning using a key identifier that is associated with the disease ``anthracnose.''~Because of this unique key identifier, the model retrieves \texttt{\textcolor{BrickRed}{``nbd''}} identifier and combines anthracnose disease with other variations, such as curled leaves, different background, or environmental conditions.
\begin{table}
  \centering
  \scriptsize 
  \begin{tabularx}{1.03\linewidth}{X c l}
    \toprule
    Hyperparameter & Values & Function \\
    \midrule
    \multicolumn{3}{c}{\textbf{Training stage}} \\
    \midrule
    Image resolution & 1024$\times$1024 & For high-quality images \\ 
    Batch size & 1 & VRAM size \\
    Gradient accumulation steps & 4 & Stabilizes training \\ 
    Gradient checkpointing & $\checkmark$ & Reduces memory usage \\
    Learning rate (LR) & 1$e$\textsuperscript{-4} & Safe value for stable convergence \\
    SNR gamma\textsuperscript{$\dagger$} & 5 & Reduces overfitting \\
    Text encoder\textsuperscript{$\ddagger$} & 5$e$\textsuperscript{-6} & Better generalizability \\ 
    Max. sequence length\textsuperscript{$\ddagger$} & 100 & For longer prompts \& detailing \\ 
    LR scheduler & constant & For limited training steps \\
    LR warmup steps & 10 & Optimized jumps for pretrained \\
    Mixed precision & fp16 & Speeds up training \\ 
    Optimizer & 8-bit Adam & Memory efficient \\
    Training steps & 2000 & Good for fine-tuning \\ 
    \bottomrule
    \multicolumn{3}{c}{\textbf{Inference stage}} \\
    \midrule
  Validation inference steps & 50 & To generate realistic images \\
  Guidance scale & 2.5 &  To adhere to prompts  \\
  \bottomrule
  \end{tabularx}
  \caption{List of all the hyperparameter fine-tuning values for training the generative models.~\textbf{\textit{Note:}}~$\dagger$ was used only in SDXL model, while $\ddagger$ was used for both SD3.5M \& L models.}
  \label{tab2}
 \end{table}
Additionally, the numerical weight \texttt{\textcolor{BrickRed}{(1.2)}} assigned to `anthracnose' in the prompt acts as a strength modifier, instructing the model to emphasize anthracnose-related features more prominently in the generated images while maintaining other variations.

\subsection{Model evaluation: Training \& Inference}
To report novel benchmarks for the models, we conducted a thorough evaluation across two key aspects: (a) computational demands, and (b) performance metrics. Under computational resources, we measured GPU memory (MiB), power consumption (W),, processing time (hours), and total energy usage (kWh). To systematically track these computational metrics during training and inference tasks, we implemented a real-time GPU monitoring script. This multi-threaded Python script logged hardware metrics in parallel with model execution, thereby extracting GPU data at 1-second intervals via the \href{https://docs.nvidia.com/deploy/nvidia-smi/}{\textcolor{blue}{NVIDIA System Management Interface)}} command. The recorded data was then stored in a structured \textit{*.csv} file for further analysis.

We also evaluated Learned Perceptual Image Patch Similarity (LPIPS) score to compare the quality of in-field and generated images~\cite{zhang2018unreasonable}. LPIPS was chosen over Peak Signal-to-Noise Ratio and Structural Similarity Index due to it close alignment with human visual perception~\cite{xing2023you}.~Unlike traditional metrics, LPIPS leverages deep learning to assess perceptual similarity.~It follows three key steps: (a) extracting features from multiple layers of a pre-trained network~\textit{(AlexNet in our case)} to capture high-level textural and structural details, (b) computing the Euclidean (L2) distance between the extracted features, and (c) weighting the feature distance to produce a similarity score between 0-1, where a lower score indicates higher perceptual similarity.

%% file: sec/5_results.tex
\section{Experimental results}
\label{results}
\subsection{Memory consumption vs. power usage}

In agriculture, memory consumption and power usage are one of the key factors when adopting models for specific applications~\cite{rai2024agricultural}. Extension educators or agricultural technology providers seek to stay updated about SoTA technologies since it can help them to improve their crop production~\cite{xu2023scoping}.~Farmers, however, focus more on benchmark metrics rather than technical complexities when adopting new technologies. Therefore, we have provided a novel benchmarking that evaluates the computational costs of GMs for large-scale synthetic crop disease images, reporting memory consumption and power usage across for all variants. 

\begin{figure*}[h]
  \centering
  \begin{subfigure}{0.32\textwidth}
    \centering
    \includegraphics[width=\linewidth]{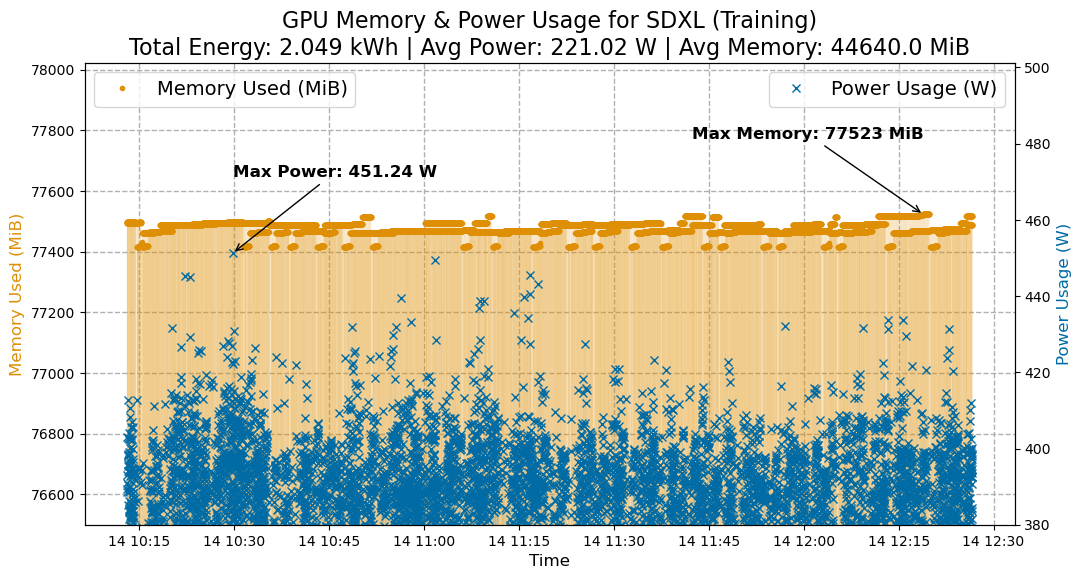}
    \caption{SDXL during training}
  \end{subfigure}
  \hfill
  \begin{subfigure}{0.32\textwidth}
    \centering
    \includegraphics[width=\linewidth]{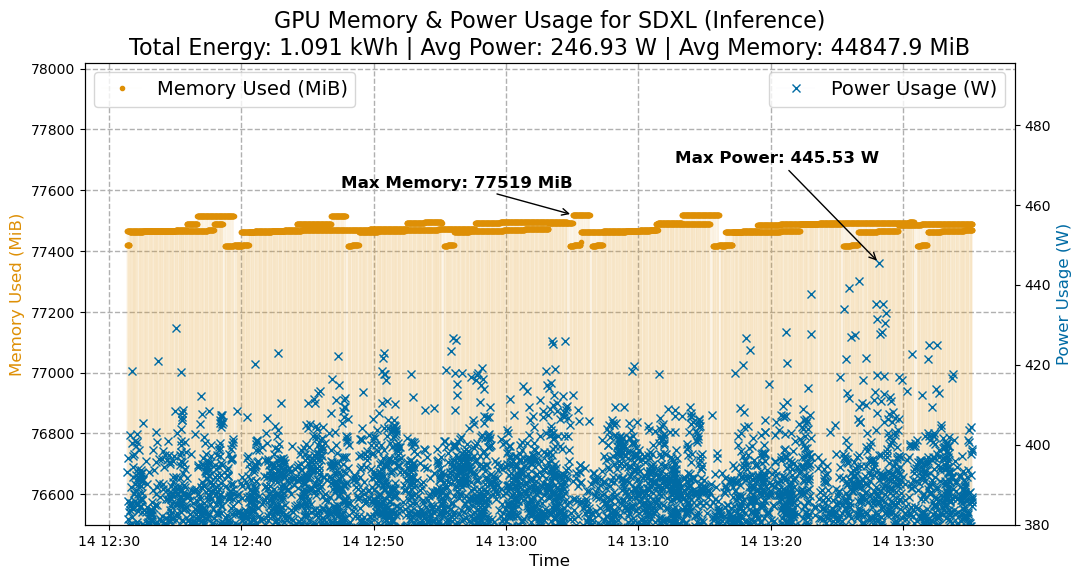}
    \caption{SDXL during inferencing}
  \end{subfigure}
  \hfill
   \begin{subfigure}{0.32\textwidth}
    \centering
    \includegraphics[width=\linewidth]{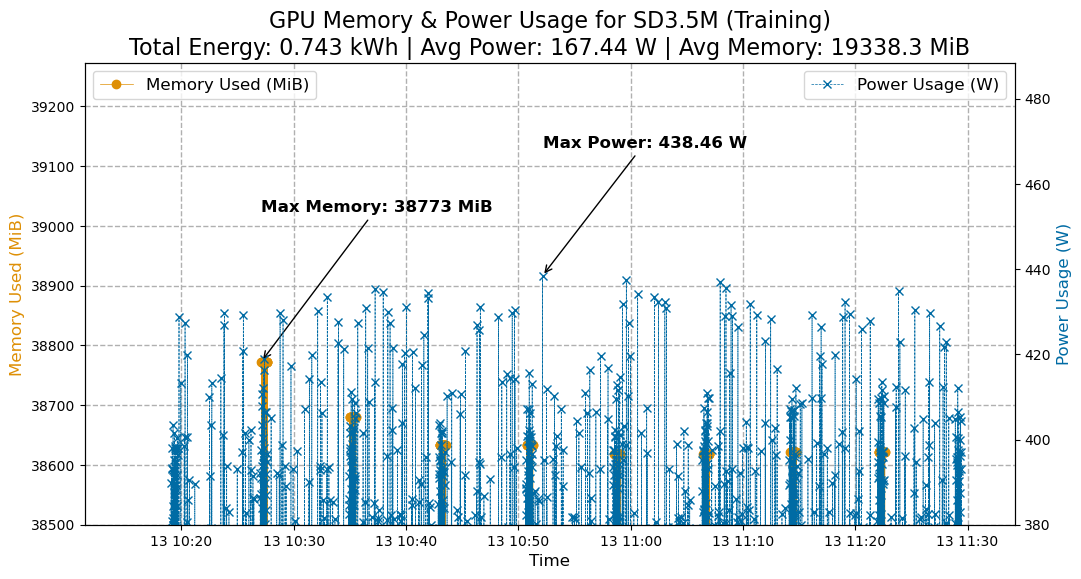}
    \caption{SD3.5M during training}
  \end{subfigure}
  \hfill
  \begin{subfigure}{0.32\textwidth}
    \centering
    \includegraphics[width=\linewidth]{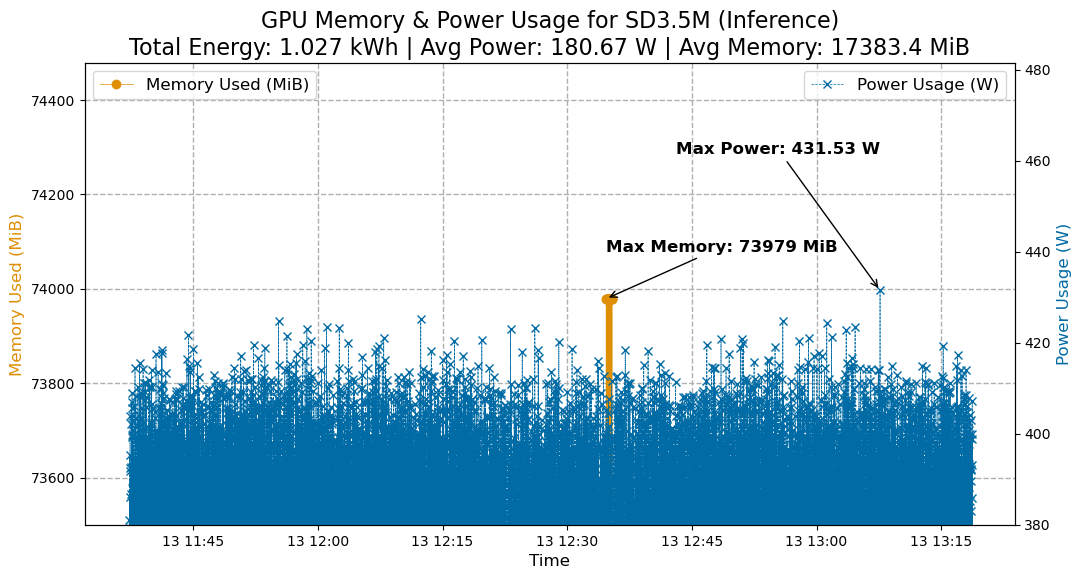}
    \caption{SD3.5M during inferencing}
  \end{subfigure}
  \hfill
\begin{subfigure}{0.32\textwidth}
    \centering
    \includegraphics[width=\linewidth]{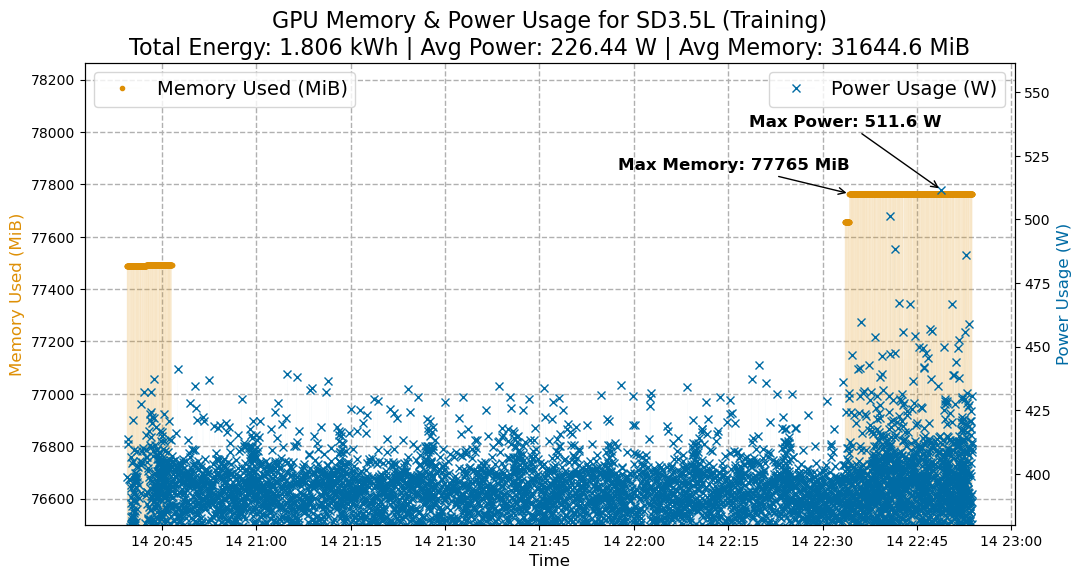}
    \caption{SD3.5L during training}
  \end{subfigure}
\hfill
  \begin{subfigure}{0.32\textwidth}
    \centering
    \includegraphics[width=\linewidth]{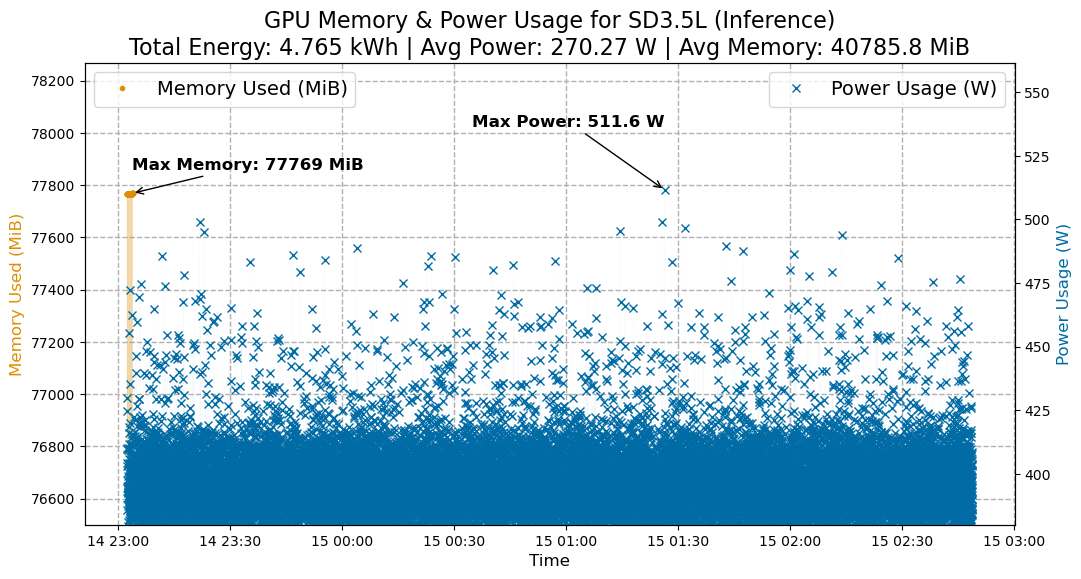}
    \caption{SD3.5L during inferencing}
  \end{subfigure}
  \caption{Comparison of \textcolor{YellowOrange}{\textbf{memory usage}} and \textcolor{NavyBlue}{\textbf{power consumption}} for SDXL, SD3.5M, and SD3.5L models during training and inference stages. The six graphs illustrate variations in resource utilization, with memory usage and power consumption plotted on the y-axes for training and inferencing stages.~\textit{\textbf{Note:} The above graphs report metrics based on generating 500 image samples.}}
  \label{memorypower}
\end{figure*}

The computational benchmarking of all the three variants shows notable differences in memory consumption and power usage (Fig.~\ref{memorypower} (a-f)).~For a fair comparison, we generated 500 disease images using each variant.~The computational results were then evaluated based on these 500 images.~Among all variants, \textit{SD3.5M is the most efficient, outperforming the rest}.~During training, SD3.5M required only 19,338 MiB ($\approx$ 20.3 GB) of memory, whereas SDXL and SD3.5L demanded 44,640 MiB ($\approx$ 46.8 GB) and 31,644 MiB ($\approx$ 33.2 GB), respectively (Fig.~\ref{memorypower}a, c, e). This translates to 2.3$\times$ lower memory when compared to SDXL variant.~In terms of power efficiency, SD3.5M consumes only 167.4 W of average power, which is 1.3$\times$ lower than SDXL (221 W) and 1.35$\times$ for SD3.5L (226.4 W). In terms of power efficiency, SD3.5M consumes only 0.7 kWh/500 images which is $\approx$ 3$\times$ lower than SDXL (2.1 kWh/500 images) and SD3.5L (1.8 kWh/500 images).~During inference (image generation) task, SD3.5M maintained its computational advantage with 17,383 ($\approx$ 18.2 GB) in memory consumption which was significantly lower than that of SDXL (47 GB) and SD3.5L (42.7 GB) (Fig.~\ref{memorypower}b, d, f).~The average power consumption during image generation task also remained balanced for the SD3.5M model which was 1.4$\times$ and 1.5$\times$ lower than SDXL and SD3.5L, respectively.~Energy consumption follows the same trend, with SD3.5M requiring only 1.1 kWh/500 images (or 0.002/image), whereas SD3.5L demands 4.7 kWh (0.009/image), making it nearly~$\approx$ 5$\times$ less efficient. 
\subsection{Time taken vs. LPIPS score}
\begin{figure}[htbp]
  \centering
    \includegraphics[width=1\columnwidth]{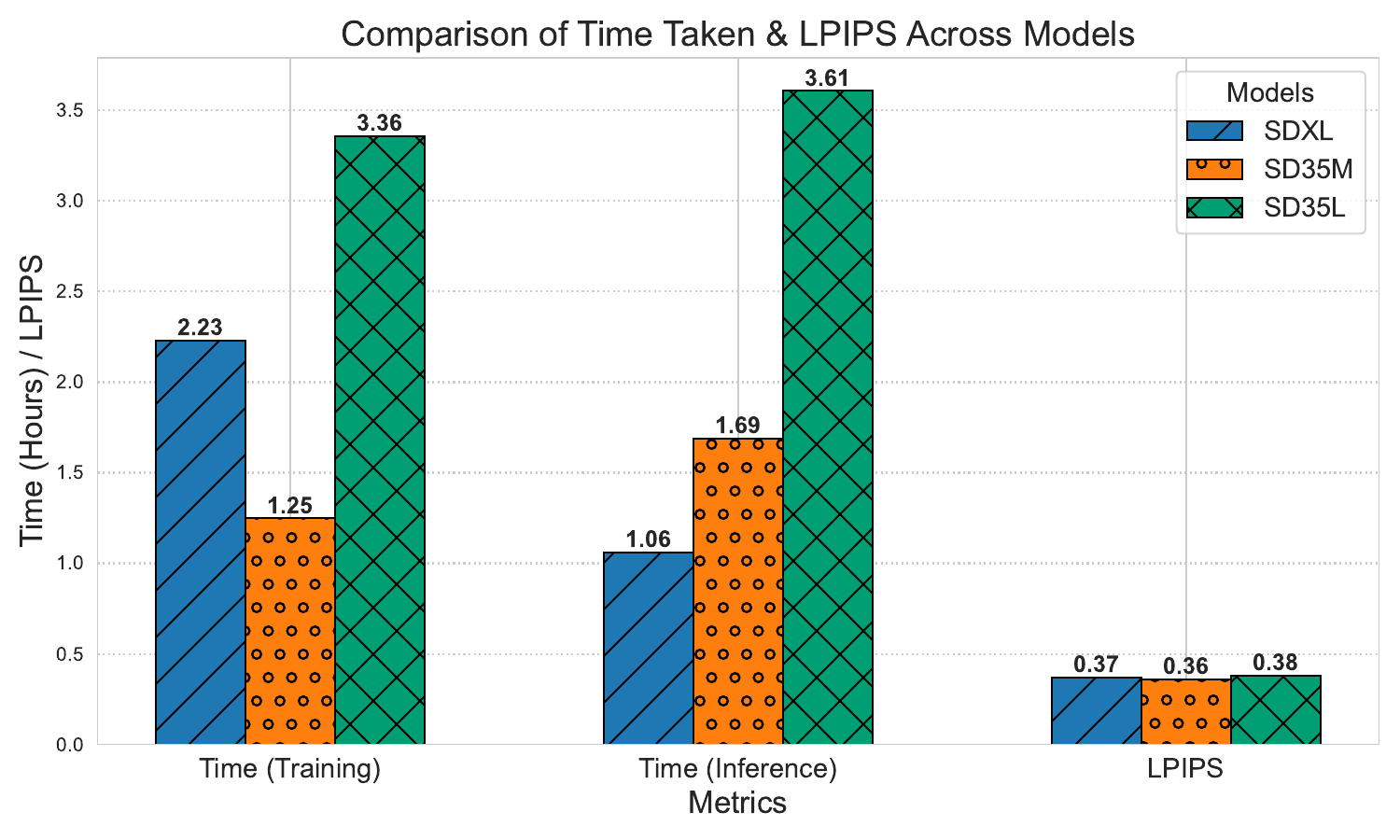}
    \vspace{-7.5mm}
    \caption{Time taken (hours) vs. LPIPS score measured for all the three variants of a multi-modal generative models.}
    \label{timevslpips}
\end{figure}

The comparison of time and energy consumption across different GMs highlight significant differences in time taken vs.~LPIPS metrics (Fig.~\ref{timevslpips}).~\textit{The SD3.5M model demonstrated the highest efficiency}, requiring only 1.25 hours for training and which is nearly half the time required by SDXL (2.23 hours).~However, the inference time for SD3.5M was 1.5$\times$ more than SDXL (1.06 hours) and $\approx$ 2.2$\times$ faster compared to SD3.5L.~Despite its longer inference time, SD3.5M achieved a lower LPIPS score (0.34), indicating its ability to generate more realistic images.~Although the LPIPS difference between SD3.5M and SDXL was only 0.01, this variation becomes significant when considering disease fine-grained symptoms.~Moreover, this 0.01 difference scales with the number of generated images, further impacting the overall dataset quality.~\textit{These results suggest that SD3.5M is the most optimal model in terms of computational efficiency, making it a preferable choice for resource-constrained applications in small-scale organizations}.~With SD3.5M success, it is also evident that training models with more parameters does not significantly increase its performance or generalization ability~\cite{hoffmann2022training}.

%% file: sec/6_discussion.tex
\section{Discussion, limitations, and future work} \label{discussion}

\begin{figure*}[!h]
    \includegraphics[scale=0.53]{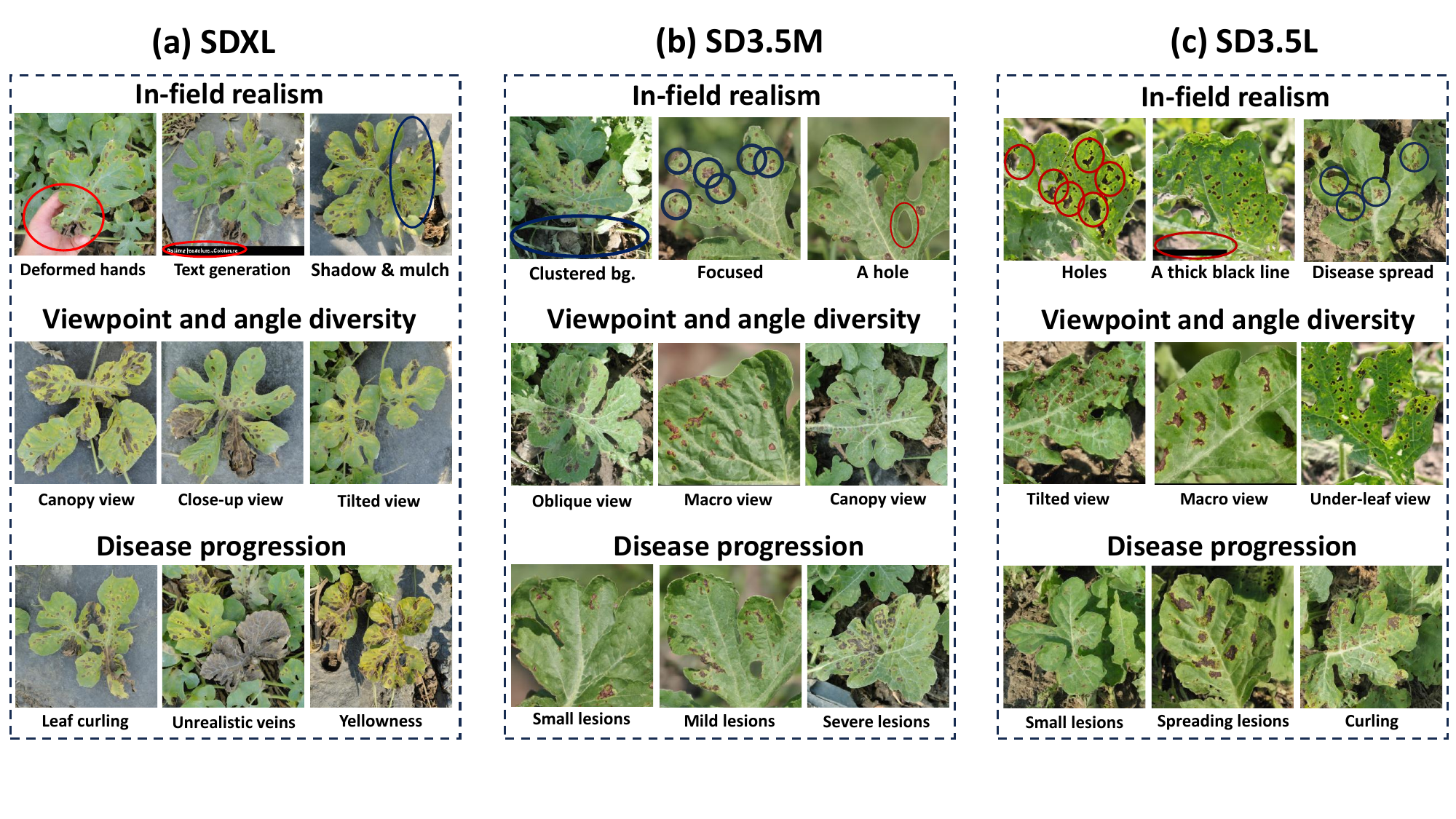}
    \vspace{-15mm}
    \caption{An example set of synthetic images generated for the watermelon disease dataset using: (a) SDXL, (b) SD3.5M, and (c) SD3.5L models. \textbf{\textit{Note:}} \textbf{\textcolor{Blue}{Dark blue}} highlights attributes where the trained model generated better-quality images, while \textbf{\textcolor{Red}{red}} indicates poor examples.}
    \label{figgeneratedimages}
  \end{figure*}

Based on the models trained to generate disease samples, Fig.~\ref{figgeneratedimages} shows the quality and diversity of synthetic images. The overall success of GM in generating a diverse synthetic crop disease depends on three important factors: (a) in-field realism with lesser artifacts, (b) viewpoint and angle diversity, and (c) progression of diseases~\cite{tariang2024synthetic}.~\textbf{In-field realism with lesser artifacts:}~For GMs to succeed, it must generate realistic images with minimal artifacts. As shown in Fig.~\ref{figgeneratedimages}a, the SDXL model generated several unexpected artifacts, such as deformed hands and unwanted  text.~Similarly, SD3.5L produced high-quality images but failed to accurately depict disease symptoms, often confusing diseases with holes or thick black lines (Fig.~\ref{figgeneratedimages}c).~In contrast, SD3.5M demonstrated the highest in-field realism, accurately representing disease symptoms (Fig.~\ref{figgeneratedimages}b).~\textbf{Viewpoint and angle diversity:}~Viewpoint and angle diversity refers to generating synthetic images from multiple perspectives rather than capturing leaves from single view.~This is a critical aspect because robotic technologies equipped with real-time sensing cameras analyze frames under varied conditions rather than a single view.~Both models,~except SD3.5L, generated canopy view (Fig.~\ref{figgeneratedimages}a, b). Moreover, we found that SD3.5L created multiple samples from an under-leaf perspective, where disease symptoms were distinctly visible (Fig.~\ref{figgeneratedimages}).
\textbf{Disease progression:}~The most challenging aspect of collecting in-field datasets for crop diseases is the extensive time required to acquire progressive disease images. Research has demonstrated that computer vision models trained on progressive disease datasets yield better results for target spraying applications using robotic sprayers~\cite{ferentinos2018deep, feng2020multi, zhang2024ismsfuse}.~Based on the images generated in this research, SD3.5M is the best-suited model to generate disease datasets with improved progression.~However, researchers should be cautious about not mixing multiple symptoms, as the model may fuse multiple lesion symptoms and generate unrealistic images (Sec.~\ref{sec:proposeddata}).

Based on this research, we also provide some critical observations (as limitations) and suggestions that could be adopted while training SD models to generate high-quality synthetic images in an agricultural scenario.~\textbf{Computational resources could be demanding:}~Although diffusion-based generative models have the potential to generate high-quality synthetic images with variations, it can be computationally demanding when it comes to training task.~As we discussed (Sec.~\ref{overview setup}), the present study utilized a single A100 GPU for training and image generation. However, the overall training and image generation time could be reduced with parallel processing.~Li et al.~\cite{li2024distrifusion}, demonstrated that using multiple GPUs is possible and we will explore this approach as part of future research.~\textbf{Mixing multiple disease symptoms can lead to unrealism:} During the analysis, we found that mixing two or more types of disease symptoms can confuse diffusion models. This will lead to the generation of unrealistic images with no specific focus on disease lesions or symptoms. Since the model is only good enough to be trained on a specific class of disease image set, it may slow down the overall process of synthetic data generation. To address this, researchers are advised to train only one disease with one symptom to generate high-quality, realistic looking samples.~\textbf{Prompt engineering can be challenging:} Current SD models use the Word2Vec~\cite{mikolov2013efficient} approach to convert billions of words into vector representations (Sec.~\ref{modelexplain}).~However, Word2Vec is not well-equipped to handle ``agricultural-centered'' terms and plant disease terminology.~While words like wilt, disease, or lesions might produce suitable results, more specific scientific prompts,~\texttt{\textcolor{BrickRed}{``Colletotrichum orbiculare''}}, often fail to generate accurate synthetic samples.~To address this, researchers are encouraged to explore AgriBERT~\cite{rezayi2022agribert} or SciBERT~\cite{beltagy2019scibert}.

%% file: sec/7_conclusions.tex
\section{Conclusion} \label{conclude}

In this research, we introduced novel benchmarking and prompt engineering approaches to generate high-quality synthetic images of common watermelon diseases.~Although the research presented focused on a specific type of disease, the same strategies can be applied to generate synthetic samples for other crop diseases.~Based on the computational metrics and performance score presented, we conclude that \textit{SD3.5M is best suited for training and generating disease samples in an agricultural scenario}.~To create realistic samples, it is essential to incorporate techniques such as Dreambooth and LoRA to help GMs to learn class-specific disease features from limited samples.~Furthermore, \textit{SD3.5M can generate over 500 synthetic images from just 36 real samples in $\approx$ 1.5 hours, offering greater variation and significantly faster processing than other SoTA models}.~This makes it cost-efficient solution for small organizations with limited access to HPC.~Moreover, the ability to rapidly generate synthetic disease samples allows AI models to quickly adapt to merging on-farm threats, thereby improving decision-making in targeted spraying applications.~Finally, establishing a standardized benchmarking framework will enable researchers to make informed decisions based on model's efficiency and lower computational demands. We recommend SD3.5M as an optimal choice for researchers working in resource-constrained environments, particularly in small organizations, to generate high-quality synthetic crop disease images in less time.

%% file: sec/8_ackn.tex
\section{Acknowledgment}

This research was supported by the US Department of Agriculture (USDA)-Small Business Innovation Research \& Technology Transfer Programs (SBIR/STTR) grant~\#~2024-51402-42007.~Thanks to Emily Witt \& Michael Sweat for their assistance with field experiments.~\textbf{This paper will appear in:~Proceedings of the IEEE/CVF Conference on Computer Vision and Pattern Recognition Workshops (CVPRW), 2025.}